\documentclass[conference]{IEEEtran}
\IEEEoverridecommandlockouts
\usepackage{cite}
\usepackage{amsmath,amssymb,amsfonts}
\usepackage{algorithm}
\usepackage{algorithmic}
\usepackage{graphicx}
\usepackage{xcolor}
\usepackage{caption}
\usepackage{subcaption}
\usepackage{tabularx}

\newcommand{\etal}{\textit{et al}. }
\newcommand{\ie}{\textit{i}.\textit{e}. }

\newcommand{\YE}[1]{{\textcolor{blue}{[#1]}}}

\usepackage{comment}
\usepackage{shortbold}
\usepackage[misc]{ifsym}

\begin{document}

\title{BayesFT: Bayesian Optimization for Fault Tolerant Neural Network Architecture}

\author{\IEEEauthorblockN{Nanyang Ye}
\IEEEauthorblockA{
\textit{Shanghai Jiao Tong University}\\
Shanghai, China \\
ynylincoln@sjtu.edu.cn}
\and
\IEEEauthorblockN{Jingbiao Mei}
\IEEEauthorblockA{
\textit{University of Cambridge}\\
Cambridge, United Kingdom \\
jm2245@cam.ac.uk}
\and
\IEEEauthorblockN{Zhicheng Fang}
\IEEEauthorblockA{
\textit{Shanghai Jiao Tong University}\\
Shanghai, China \\
fangzhicheng@sjtu.edu.cn}
\and
\IEEEauthorblockN{Yuwen Zhang}
\IEEEauthorblockA{
\textit{University College London}\\
London, United Kingdom \\
yuwen.zhang.20@ucl.ac.uk}
\and
\IEEEauthorblockN{Ziqing Zhang}
\IEEEauthorblockA{
\textit{University of Cambridge}\\
Cambridge, United Kingdom \\
zz404@cam.ac.uk}
\and
\IEEEauthorblockN{Huaying Wu}
\IEEEauthorblockA{
\textit{Shanghai Jiao Tong University}\\
Shanghai, China \\
wuhuaying@sjtu.edu.cn}
\and
\IEEEauthorblockN{Xiaoyao Liang}
\IEEEauthorblockA{
\textit{Shanghai Jiao Tong University}\\
Shanghai, China \\
liang-xy@sjtu.edu.cn}
}

\maketitle

\begin{abstract}
To deploy deep learning algorithms on resource-limited scenarios, an emerging device-resistive random access memory (ReRAM) has been regarded as promising via analog computing. However, the practicability of ReRAM is primarily limited due to the weight drifting of ReRAM neural networks due to multi-factor reasons, including manufacturing, thermal noises, and etc. In this paper, we propose a novel Bayesian optimization method for fault tolerant neural network architecture (BayesFT). For neural architecture search space design, instead of conducting neural architecture search on the whole feasible neural architecture search space, we first systematically explore the weight drifting tolerance of different neural network components, such as dropout, normalization, number of layers, and activation functions in which dropout is found to be able to improve the neural network robustness to weight drifting. Based on our analysis, we propose an efficient search space by only searching for dropout rates for each layer. Then, we use Bayesian optimization to search for the optimal neural architecture robust to weight drifting. Empirical experiments demonstrate that our algorithmic framework has outperformed the state-of-the-art methods by up to 10 times on various tasks, such as image classification and object detection.
\end{abstract}

%\begin{IEEEkeywords}
%ReRAM, Deep Neural Networks
%\end{IEEEkeywords}

\section{Introduction}
Deep learning has achieved tremendous success in various fields, such as image classification, objection detection, natural language processing, and autonomous driving. To deploy deep learning algorithms on resource limited scenarios, such as internet of things, a lot of research has been conducted on integrating deep learning algorithms into deep neural network (DNN) accelerators, such as FPGAs, and domain specific ASICs. Whereas these approaches have demonstrated energy, latency and throughout efficiency improvements over traditional ways of using a general-purpose graphic computing unit (GPU), one inherent limitation is that digital circuits consume a lot of power to maintain high enough triggering voltage to differentiate two states. Besides, unlike human brains where neurons are all capable of computation and storage, information has to be transmitted repeatedly between computing component and memory to update DNNs. These properties are fundamentally different from human brains and lead to high energy costs and arguably deviating our DNN systems from emulating human intelligence.

To build machines like humans, neuromorphic computing has been proposed to simulate the human brain circuits for deep learning, which receives wide attention both from academia and industry. One emerging trend in neuromorphic computing is resistive random access memory (ReRAM) for deep learning with memristors \cite{shafiee2016isaac,Liu2018,Yao2020Fully}. Memristor is a non-volatile electronic memory device and one of the four fundamental electronic circuit components taking decades to be realized.

%With memristors, a common operation in DNN---in-place dot product can be implemented efficiently via analog computing without the need to move data from memory to computing device. This makes ReRAM behave more similar to human brains compared with other DNN accelerators.

However, ReRAM has been demonstrated to be not well compatible with existing deep learning paradigms designed for deterministic circuit behaviors. Due to the analog property of ReRAM, the stability of ReRAM can be largely affected by thermal noises, electrical noises, process variations, and programming errors. The weights of DNN represented by the memristance of a memristor cell, can be easily distorted, largely jeopardizing the utility of the ReRAM deep learning systems. 

To mitigate the negative effects of memristance distortion, several methods have been proposed whereas most of the settings are at the cost of extra hardware costs. For example, Liu \etal first learned the importance of neural network weights and then finetuned the important weights that were distorted \cite{2017Rescuing}. Chen \etal proposed a method to re-write DNN into ReRAM after diagnosis for each ReRAM device. This approach is not scalable as re-training DNN is needed for each weight distortion pattern of ReRAM devices \cite{chen2017accelerator}. While improvements have been observed, these methods ignore factors, such as programming errors and weight drifting during usage. Besides, they are not scalable for massive production of ReRAM devices. Diagnosing and re-training DNNs for each ReRAM device are time-consuming and expensive. Recently, Liu \etal mitigated this problem with a new DNN architecture by substituting the error correction code scheme of the commonly-used softmax layer for outputting the prediction for image classification tasks \cite{Liu2019}. In this approach, instead of predicting each class's probability in image classification, it computed a series of binary codes from images and predicted the image's class by comparing the series of codes with each class's codes precomputed and stored in a codebook. For example, if for an image, the computed code is 10001, and in the codebook, the class cat's code is 10000 and the dog's code is 11111. As the computed code has a smaller hamming distance to the class cat's code, the neural network will output cat as the result. Although this method does not need re-training DNNs each time, in this scheme, the errors caused by the previous layer's weight drifts will propagate to later layers, leading to high error entanglement in the last layers responsible for generating codes. Besides, the error-correction scheme is designed for image classification tasks cannot be directly implemented in object detection tasks that are crucial in many applications, such as autonomous driving. 

\begin{comment}
\begin{figure}[!ht]
\includegraphics[width=0.8\linewidth]{BayesFT_framework.pdf}
    \caption{Algorithmic framework of BayesFT.}
\end{figure}
\end{comment}

%\YE{should we use past tense in the next two paragraphs when presenting our work and contribution? Currently only the last sentence is in past tense.}
In this paper, we revisited the problem of fault tolerance of neural networks and identified several factors that are crucial to the robustness to weight drifting. Perhaps surprisingly, we found that the architectural choice (\ie dropout, normalization, and complexity of models, etc) played an essential role in determining the robustness to weight drifting. We proposed a Bayesian optimization method to automatically search for fault tolerant neural network architectures. We name this approach \emph{\textbf{``BayesFT''}}.

Our contributions can be summarized as follows:
\begin{enumerate}
    \item We systematically analyzed the weight drifting robustness of different neural architecture components. We identified key architectural factors in determining the weight drifting robustness, such as dropout, normalization, and complexity of models.
    \item Based on our analysis, we proposed a Bayesian optimization framework---BayesFT to automatically search for the fault tolerant neural network architectures that are robust to weight drifting in ReRAM devices.
    \item We conducted extensive numerical experiments on various tasks and datasets, such as CIFAR-10, traffic sign recognition for image classification and PennFudanPed for object detection. Results demonstrated that our methods could improve robustness by more than 10-100 times with only negligible computational costs and engineering efforts. 
\end{enumerate}
%We will release the source codes of the paper upon the acceptance.
 
\section{Preliminary}
\subsection{Basics of DNN}
A DNN can be viewed as the composition of many non-linear functions. Formally, given input data $\xB \in \Rcal^{d}$ and its corresponding label $\yB$, the task is to minimize the loss $\ell(f_{\thetaB}(\xB), \yB)$, where $\ell$ is the loss function, $f$ is the neural network with weights $\wB$. For a $K$ layer neural network, $f$ can be viewed as the composite of a series of functions $f=f_{1}\circ f_{2} \circ \cdots \circ f_{K}$. There are several commonly used layers in DNN. Convolutional layers extract features with convolution operation based on learned kernels. Fully connected layers apply non-linear function after matrix product. For more detailed introductions, we refer readers to the deep learning book \cite{Goodfellow-et-al-2016}.

\begin{comment}
Next, we will give brief introduction to two layers that might be less frequently formally investigated in this line of research.
\paragraph{\textbf{Dropout layers}} Dropout layers randomly remove neurons \YE{delete 'out'}out with a probability \YE{delete 'that is'}that is referred to as the dropout rate, which \YE{has proven effective in mitigating}is found to be effective in reducing over-fitting in machine learning research \cite{Srivastava2014}. More formally, given a weight matrix $\thetaB$, the dropout layer applies a binomial distribution $\IB \sim \text{Bernoulli}(p)$: $\thetaB \leftarrow \thetaB \IB$, where $p$ is the dropout rate.

\paragraph{\textbf{Batch normalization layers}} To constrain the output value of each layer to a reasonable range for numeric computation and avoid gradient explosion, batch normalization normalizes the input features of each data batch with learned coefficients: $\zB \leftarrow \gamma*(\xB-\mathbb{E}(\zB))/(\text{Var}(\zB)+\epsilon)+\beta$, where $\mathbb{E}(\zB)$, $\text{Var}(\zB)$ are the mean and variance of features, $\epsilon$ is a small constant to avoid zero denominator. $\gamma$ and $\beta$ are two learned coefficients \cite{IoffeS15}.
\end{comment}

\subsection{Memristance drifting modeling}
Following the setting of \cite{Liu2019} and \cite{chen2017accelerator}, to simulate the memristance drifting due to multi factors as mentioned above, we apply the following drifting term to each neural network weight $w$:
\begin{equation}
    \theta' \leftarrow \theta e^{\lambda},\quad \lambda \sim \mathcal{N}(0,\sigma^2)
    \label{eq:var}
\end{equation}
where $\theta'$ is the drifted neural network parameters, which follows a log-normal distribution. We can vary $\sigma$ to change the level of variation to simulate different ReRAM devices and deployment scenarios. It is worth noting that although we consider this setting in our paper, our methodology can be seamlessly extended to other possible weight drifting distributions.

%Various work has been done to solve this problem. Such as the Fault-Tolerant Neural Network Architecture (FTNA)\cite{Liu2019} which introduces a collaborative logistic classifier coupled with an optimized variable-length decode-free ECOC. They assign the significance on each classifier to show the correlations among the classifiers, as well as a regularization term to be applied on the loss function to rectify the classifier significance. They further set a pending zone in logistic regression ($[0.4, 0.6]$) to solve the issue of indecisiveness due to the rectified significance. The confusion matrix and number of classifiers are sent to the DNN-favorable searching code to create the codeword list, then the weights of collaborative logistic classifiers will be fine-tuned through transfer learning. Another work is hard-ware based \cite{Chen2017}, which proposes a cross-layer solution leveraging the inherent self-healing capability of the neural networks. Given the mapping, the weight gives the highest variation is reduced, and the model is re-trained. This process iterate until the test-rate converges. 

The weight drifting in ReRAM can cause significant performance degradation for DNNs. To visualize this, a plotting of simple binary classification dataset generated with Scikit-Learn is presented. As the level of weight perturbation increases, the shape of the decision boundary shifts and therefore reduces the accuracy of classification. These figures give the intuition that the weight perturbation would cause reduction in classification accuracy. 

\begin{figure}[htbp]
    \centering
    \includegraphics[width=0.6\linewidth]{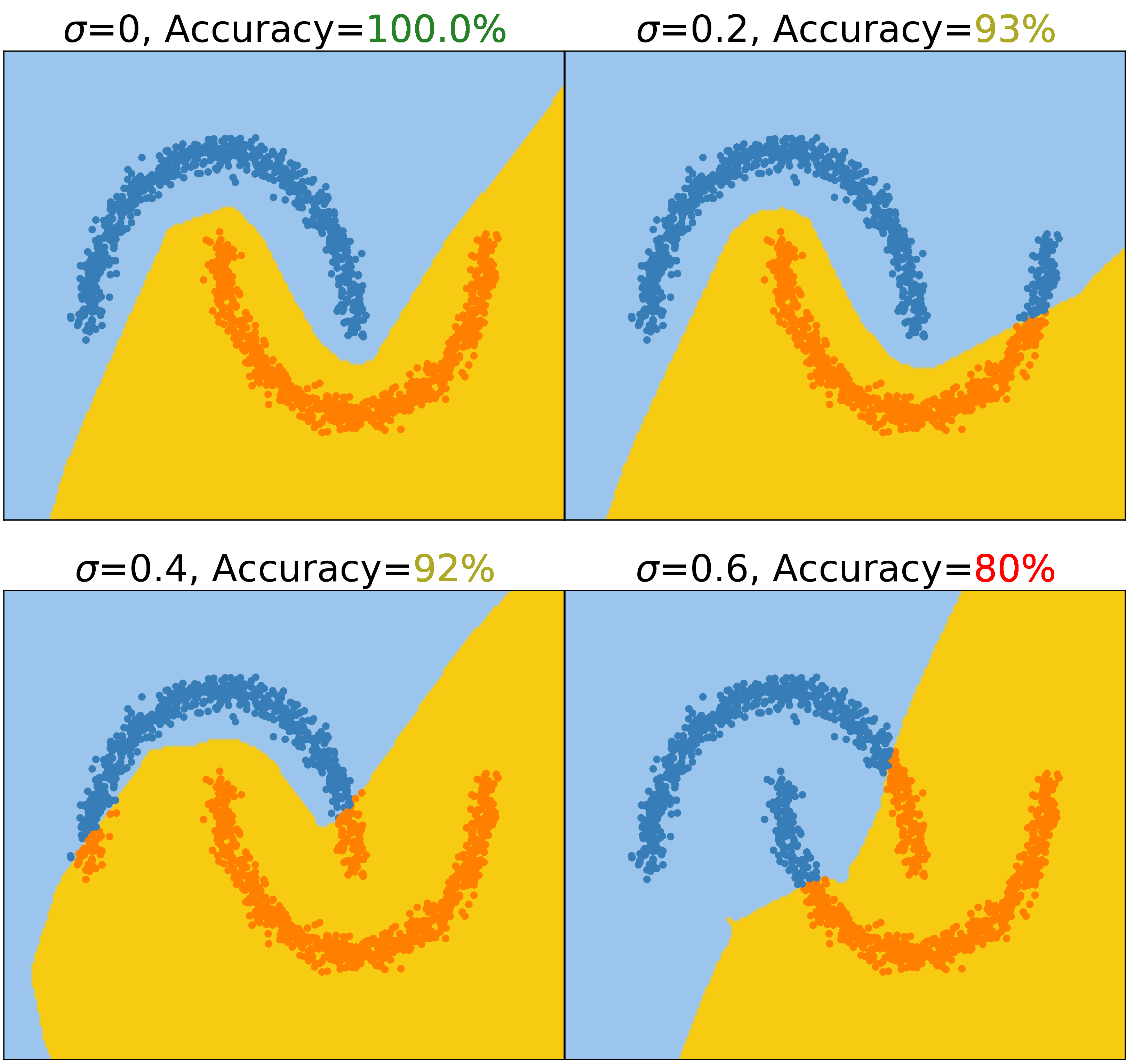}
    \caption{Decision boundary shifts caused by memristance drifting}
    \label{Binary Classifier with Weight Perturbation}
\end{figure}

\section{BayesFT: Bayesian Optimization for Fault Tolerant Neural Network Architecture}

\subsection{Exploration of fault tolerant neural architecture}
\label{sec:explore}
We first do an ablation study to investigate the fault tolerance of neural architecture factors, such as dropout, normalization, model complexity, and activation function using a multi-layer perceptron (MLP) on MNIST dataset\footnote{Same experiments are also conducted with larger models on CIFAR-10 dataset and the results are similar.}. The results are shown in Figure 2. Next, we will discuss the experiment results in detail.

\begin{figure*}[!ht]
\begin{center}
\setlength\tabcolsep{0pt}
\begin{tabular}{cccc}
\includegraphics[width=0.5\columnwidth]{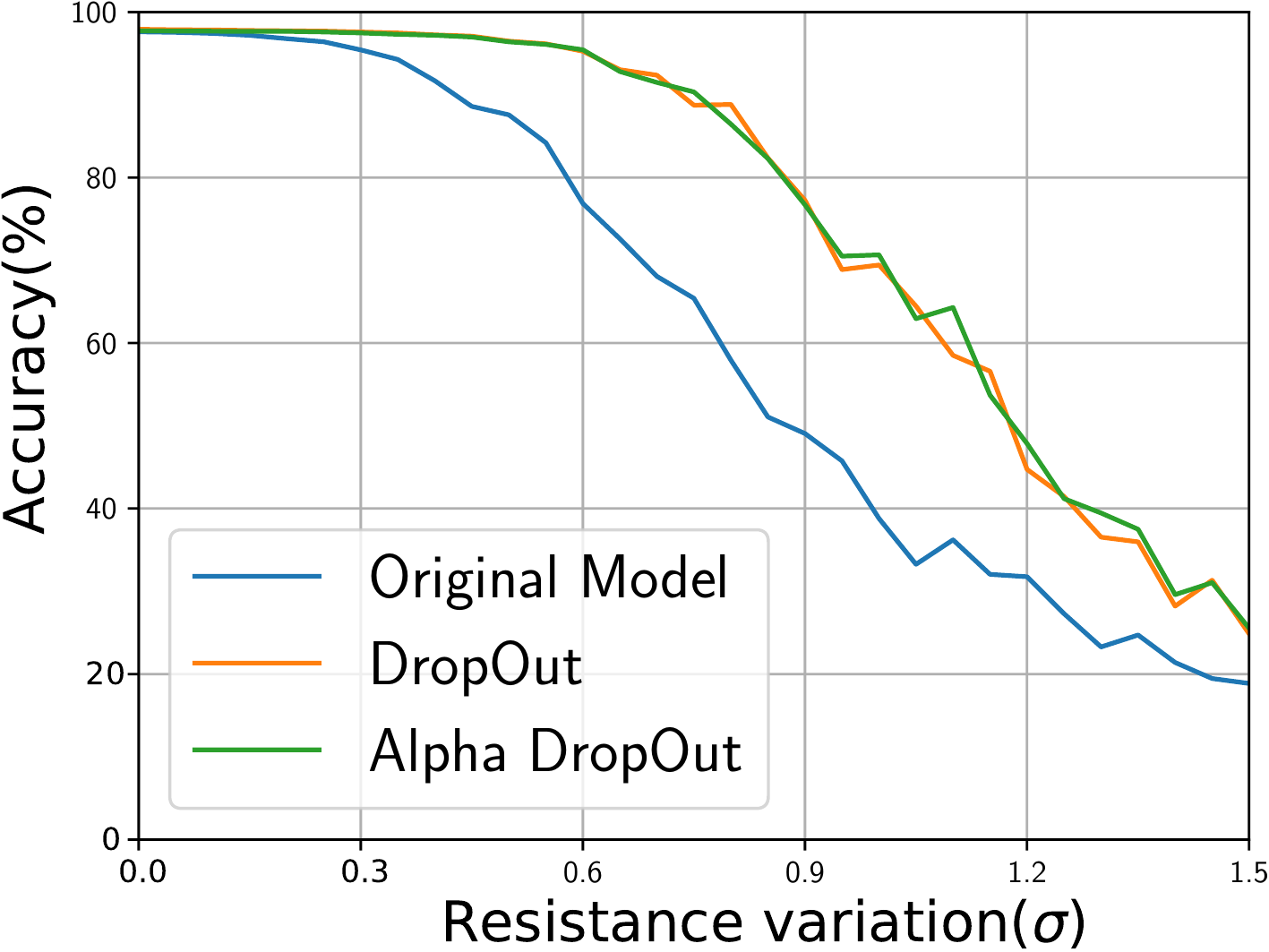}&
\includegraphics[width=0.5\columnwidth]{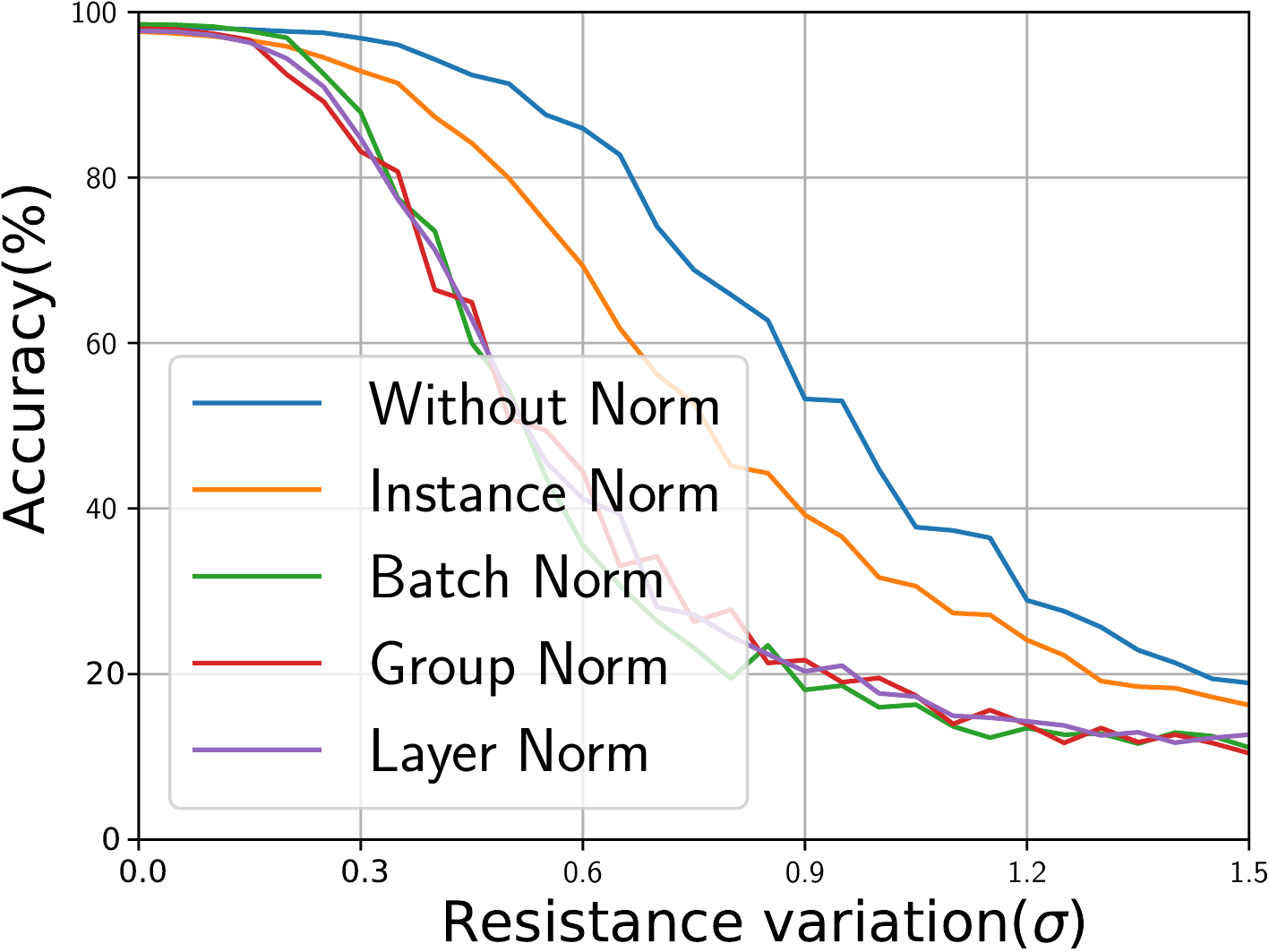}&
\includegraphics[width=0.5\columnwidth]{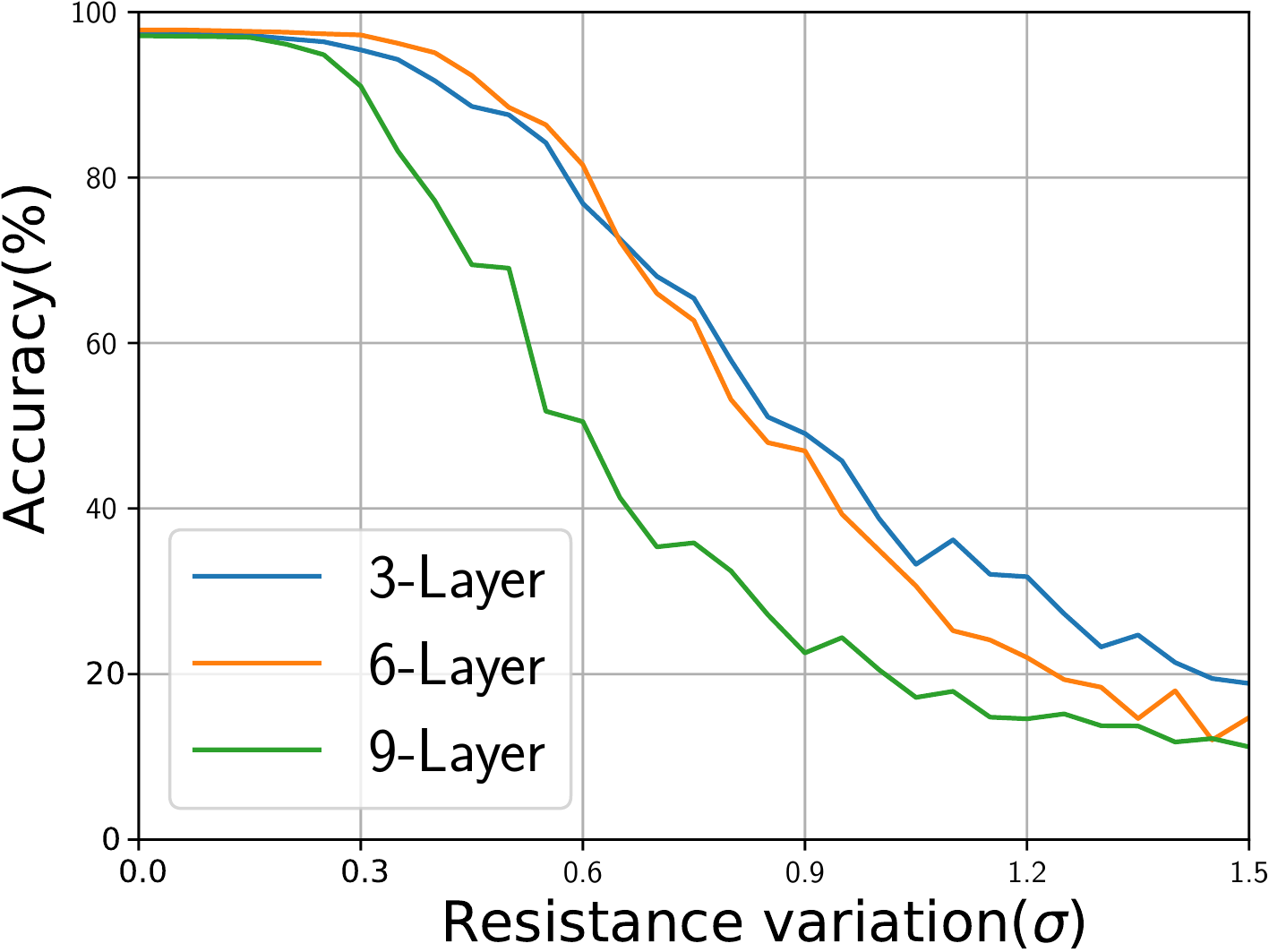}&
\includegraphics[width=0.5\columnwidth]{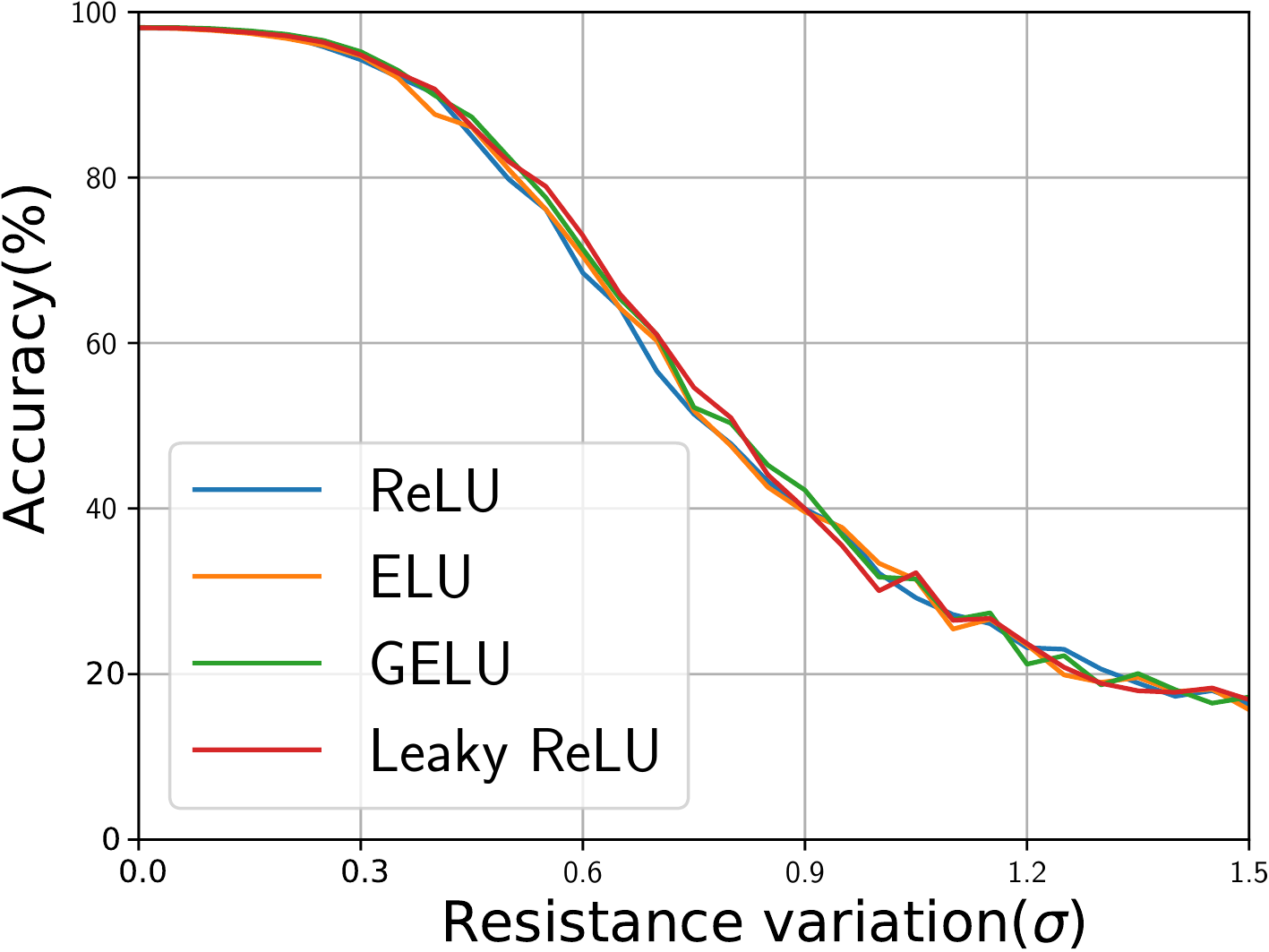}\\

(a) Dropout (*) &(b) Normalization (*) &(c) Model complexity (*) &(d)  Activation functions
\end{tabular}
\end{center}
\caption{Exploration of neural network architecture factors for fault tolerance. (*) indicates there are significant differences between methods.}
\label{fig:exploration}
\end{figure*}

\paragraph{\textbf{Dropout}}
Dropout layers randomly remove neurons with a probability, often referred to as the dropout rate, which has proven effective in mitigating over-fitting in machine learning research \cite{Srivastava2014Dropout}. More formally, given a weight matrix $\thetaB$, the dropout layer applies a binomial distribution $\IB \sim \text{Bernoulli}(p)$: $\thetaB \leftarrow \thetaB \IB$, where $p$ is the dropout rate. We also explore a variant of dropout---alpha dropout \cite{klambauer2017self}. Perhaps surprisingly, Figure~\ref{fig:exploration}(a) reveals that dropout layers significantly improve the robustness of network to memristance drifting. The network with dropout layer could gain self-healing ability while randomly removing weights during training process. The enhanced robustness to missing weights transfers to robustness to weight drifting. The performance of original dropout and alpha dropout are similar. The alpha dropout maintains the original input mean and variance by re-scaling after dropout. The re-scaling increases computational cost without improving the performance significantly compared to original dropout. We thus focus on the dropout for later studies.

\paragraph{\textbf{Normalization}}
%Experiment of BatchNormalization was carried out. Original Batchnormalization [Paper] , Group Normalization, Layer Normalization and Instance Normalization are carried out. [Paper] and explanation of these different normalization approaches are explained. %
We considered different normalization methods commonly used in machine learning, \ie batch normalization~\cite{ioffe2015batch}, layer normalization~\cite{ba2016layer}, instance normalization~\cite{ulyanov2016instance}, and group normalization~\cite{2018Group}. Each normalization method normalizes across some dimension of input features. Normalization methods empirically facilitate the convergence of optimization algorithms in machine learning. 

%\begin{figure}[H]
%    \centering
%    \includegraphics[width=0.98\linewidth]{normalizations.jpeg}
%    \caption{\textbf{Normalization methods}. Each subplot shows a feature map tensor, with $N$ as the batch axis, $C$ as the channel axis, and $(H,W)$ as the spatial axis. The pixels in blue are normalized by the same mean and variance, computed by aggregating the values of these pixels.\cite{2018Group}}
%    \label{fig:Normalization}
%\end{figure}

For this family of feature normalisation methods, we perform the following computation:
\begin{equation}
    \hat{x}_i = \beta \dfrac{x_i-\mu_i}{\sigma_i} + \gamma
\end{equation}
For $\hat{x}$, the features computed by a layer, $\mu_i$ and $\sigma_i$ are mean and variance over the set $\mathcal{S}_{i}$ respectively. $i$ is the index, e.g. for 2D images $i=(i_N, i_C, i_H, i_W)$ is a 4D vector representing the features in $(N,C,H,W)$ order, where $N$ is the batch axis, $C$ is the channel axis, $H,W$ are the spatial height and width axis. $\beta$ and $\gamma$ are parameters to be trained in normalization layers. From Figure~\ref{fig:exploration}(b), adding normalization generally worsens the performance. This is because normalizing the hidden layer's input to have zero mean and unit variance and then being linearly transformed creates ``Achilles's heel'' as weight drifting on transformation parameter $\beta$ and $\gamma$ can be more harmful because of the normalization.

%\paragraph{\textbf{Residual Connection}}%
%We examined the performance of having convolutional layers inside residual and not having convolutions layer inside residual connection. %

\paragraph{\textbf{Model complexity}}
We tested whether more complex models or deeper DNNs can improve robustness. As shown in Figure~~\ref{fig:exploration}(c), increasing model complexity \ie number of layers can decrease performance. This is because drifted weights accumulate errors as the layer goes deeper. This indicates that naively applying neural architecture search (NAS) method can lead to sub-optimal performance as most NAS search spaces are built on larger models for accuracy.

\paragraph{\textbf{Activation Function}}
The robustness of networks with ReLU activation function, leaky ReLU, ELU and GELU\cite{hendrycks2016gaussian} is also explored. From Figure~~\ref{fig:exploration}(c), there are no statistical differences between these methods.

\subsection{Bayesian optimization for fault tolerant neural network architecture}
\label{sec:BayesIntro}
From the above analysis, we found that adding dropout layers can improve the neural network's robustness to weight drifting significantly. However, misspecification of dropout layers can also lead to suboptimal performances. To automatically search for the optimal neural network architecture robust to weight drifting. To simplify the neural architecture search, instead of searching for all possible topology structures of DNNs, we append dropout layers after each DNN layer except the last softmax layer for output and search for the dropout rate of each layer only. We denote the specification of neural network architecture as $\alphaB \in \Rcal^{K-1}$, where $K$ is the number of layers of DNNs for architectural selection. In addition to its simplicity, this neural architecture search space design is also compatible with all existing neural network architectures. This makes our method suitable for integrating into many platforms where not all neural network architectures are supported. As there is no exact gradient information available for $\alphaB$, we consider Bayesian optimization to search for the optimal $\alphaB$ from the search space \cite{Snoek12practicalbayesian}. 

We first define our objective function by marginalizing the loss over the distribution of drifting neural network parameters $\thetaB$:
\begin{align}
    u(\alphaB, \thetaB) &= -\mathbb{E}_{\tilde{\thetaB} \sim p(\tilde{\thetaB})}[\ell(f_{(\alphaB, \tilde{\thetaB})}(\xB), \yB)]
\end{align}
where $\tilde{\thetaB} = \thetaB \exp^{\lambda}$ ,$\lambda \sim \Ncal(0, \sigma^2)$, $\ell(f_{(\alphaB, \tilde{\thetaB})}(\xB), \yB)$ is the loss of a neural network with architecture $\alphaB$ and parameter $\thetaB$ given input data $\xB$ and target $\yB$. This intractable equation can be approximately computed by Monte Carlo sampling:
\begin{align}
    u(\alphaB,\thetaB) &\simeq -\frac{1}{T}\sum_{t=1}^{T}\ell(f_{(\alphaB, \tilde{\thetaB}_{t})}(\xB), \yB)]
\end{align}
where $T$ is the number of Monte Carlo samples and $\tilde{\thetaB}_{t}$ is the $t$-th sample randomly drawn from $p(\tilde{\thetaB})$. For maximizing the objective function, we use an optimization scheme where $\alphaB$ and $\thetaB$ are optimized alternatively. When optimizing $\alphaB$, we use Bayesian optimization as the gradient for $\alphaB$ is not available. Bayesian optimization uses a surrogate model constructed from previous trials to determine the point for next trial,\ie the point which is the most likely to give the optimal solution for the gradient-free optimization problem \cite{brochu2010tutorial}. For $\thetaB$, we use the stochastic gradient descent method.

For Bayesian optimization, we use a Gaussian process regression model as the surrogate model. Suppose we already have $n$ trials of different settings of $\alphaB$ denoted as $\alphaB_{1:n}$, its corresponding objective function value $g(\alphaB_{1:n})$, and kernel matrix $\kappa(\alphaB_{1:n}, \alphaB_{1:n})$, more specifically:
\begin{align}
    \alphaB_{1:n}&=[\alphaB_1, \cdots, \alphaB_n]\\
    g(\alphaB_{1:n})&=
    \begin{bmatrix}
    u(\alphaB_{1}, \thetaB),& \cdots,& u(\alphaB_{n}, \thetaB)
    \end{bmatrix}\\
    \kappa(\alphaB_{1:n}, \alphaB_{1:n})&=
    \begin{bmatrix}
    \kappa(\alphaB_{1}, \alphaB_{1}),& \cdots, &\kappa(\alphaB_{1}, \alphaB_{n})\\
    \cdots, &\cdots, &\cdots \\
    \kappa(\alphaB_{n},\alphaB_{1}),&\cdots, &\kappa(\alphaB_{n}, \alphaB_{n})
    \end{bmatrix}
    %[\\
    %\kappa(\alphaB_{1:n}, \alphaB_{1:n})&=[\kappa(\alphaB_{1}, \alphaB_{1}), \cdots, \kappa(\alphaB_{1}, \alphaB_{k});\\ \cdots \\; &\kappa(\alphaB_{k},\alphaB_{1}), \cdots, \kappa(\alphaB_{k}, \alphaB_{k})]
\end{align}

Then, according to Gaussian process's property, the posterior probability of $g(\alphaB)$ after $n$ trials follows a Gaussian distribution:
\begin{align}
    &p(g(\alphaB)|g(\alphaB_{1:n})) \sim \Ncal\left(\mu_{n}(\alphaB), \sigma^2_{n}(\alphaB)\right) \\
    &\kappa_{n}(\alphaB)= \kappa(\alphaB, \alphaB_{1:n})\kappa(\alphaB_{1:n}, \alphaB_{1:n})^{-1}g(\alpha_{1:n}) \nonumber \\
    &\sigma^2_{n}(\alphaB)=\kappa(\alphaB, \alphaB)-\kappa(\alphaB, \alphaB_{1:n})\kappa(\alphaB_{1:n},\alphaB_{1:n})^{-1}\kappa(\alphaB_{1:n},\alphaB) \nonumber
\end{align}
where $\kappa$ is the kernel function. In our experiment, we use the exponential kernel function:
\begin{equation}
    \kappa(\alphaB_1, \alphaB_2) = k_{0}\exp(-||\alphaB_1-\alphaB_2||^2)
\end{equation}
where $||\alphaB_1-\alphaB_2||^2=\sum_{i=1}^d k_i(\alpha_{1,i}-\alpha_{2,i})^2$, and $k_{0:d}$ are parameters of the kernel.

Then, the next trial is given by finding the point that is most likely to result in the optimal objective value: $\alphaB=\max_{\alphaB} p(g(\alphaB)|g(\alphaB_{1:n}))$. Thus, we have the algorithm based on Bayesian optimization for fault tolerant neural network architecture as shown in Algorithm~\ref{algo:Bayes}.

\begin{algorithm}
\caption{Bayesian Optimization for Fault Tolerant DNN}
\label{algo:Bayes}

\begin{algorithmic}[1]
%\hspace*{\algorithmicindent} \textbf{Input} \\
%\hspace*{\algorithmicindent} \textbf{Output} 
 
\STATE \textbf{Input:} Dataset $(\xB,\yB)$, neural network parameters $\thetaB$, dropout rates for each layer $\alphaB$, number of epochs for training neural networks $E$. \nonumber
\STATE \textbf{Output:} Trained neural network $\thetaB$ and dropout rates for each layer $\alphaB$. \nonumber
\STATE \textbf{Initialization:} initialize $\thetaB$ with Xavier random initialization \cite{Glorot10understandingthe}, $\alphaB$ with a uniform distribution on $[0,1]$, number of iterations $t=0$:
%\begin{align}
%    &\theta_{0} \sim \text{Xavier} \\
%    &\alphaB_{0} \sim \text{uniform}[0,1] \\
%    & t \leftarrow 0
%\end{align}

\REPEAT 

\FOR{$e = 1$ to $e=E$}
\STATE Optimize neural network parameters $\thetaB$:
\begin{equation}
    \thetaB_{t} \leftarrow \thetaB_{t-1} - \nabla_{\thetaB}u(\alpha_{t-1}, \thetaB_{t-1})
    \nonumber
\end{equation}
\ENDFOR

\STATE Update the posterior distribution function for Bayesian optimization:
\begin{align}
    &g(\alphaB_{1:t-1})=
    \begin{bmatrix}
    u(\alphaB_{1}, \thetaB_{t}), \cdots,  u(\alphaB_{t-1}, \thetaB_{t})
    \end{bmatrix}\\
    &\kappa_{t-1}(\alphaB)= \kappa(\alphaB, \alphaB_{1:t-1})\kappa(\alphaB_{1:t-1}, \alphaB_{1:t-1})^{-1}g(\alpha_{1:t-1}) \nonumber \\
    &\sigma^2_{t-1}(\alphaB)=\kappa(\alphaB, \alphaB) -\kappa(\alphaB, \alphaB_{1:t-1})\\&\kappa(\alphaB_{1:t-1},\alphaB_{1:t-1})^{-1}\kappa(\alphaB_{1:t-1},\alphaB) 
\end{align}
\STATE Calculate the optimal $\alphaB$ from the updated posterior distribution function for the surrogate model: 
\begin{equation}
    \alphaB_{t} \leftarrow \max_{\alphaB} p(g(\alphaB)|g(\alphaB_{1:t-1})) \nonumber
\end{equation} 
\STATE $t \leftarrow t + 1$

\UNTIL convergence;
\end{algorithmic}
\end{algorithm}

%\subsection{\YE{Bayesian Optimization for neural network architecture search}}

%\subsection{\YE{Bayesian full neural architecture search}}
%Two phase:
%(1)NAS without training
%(2)Bayesian fine-tuning

\section{Experiment results}
In this section, we discuss the experimental details to evaluate our BayesFT. The first section describes the implementation tasks and details. And then, we discuss the illustrative results. For all our experiments, we used a server with Intel Xeon E5-2630 CPU and Geforce RTX 2080Ti GPUs. We used Pytorch 1.7.0 and CUDA 10.1 for implementing numerical experiments. We consider the experimental setting where neural networks are trained off-line on GPU servers and then deployed on ReRAM devices. This is arguably more realistic as diagnosing and correcting ReRAM on-line may largely increase latency and energy consumption. For comparison, we implemented the following baseline algorithms:
\begin{enumerate}
    \item \textbf{Empirical risk minimization(ERM)}:the baseline algorithm that only minimizes the empirical risk. 
    \item \textbf{ReRAM-variations (ReRAM-V,\cite{chen2017accelerator})}: ReRAM-V diagnoses the ReRAM circuits and readjusts the weights iteratively to improve the robustness to weight drifting until convergence.
    \item \textbf{Adversarial weight perturbations (AWP, \cite{wu2020adversarial})}: AWP conducts adversarial training on the weight perturbations of neural networks to improve the robustness against weight shifts.
    \item \textbf{Fault tolerant neural network architecture (FTNA,\cite{Liu2019})}: FTNA replaced the original model's last softmax layer with an error-correction coding scheme as discussed before. %\footnote{For implementing FTNA,  the last two layers were fixed for encodingout attempt of implementing it based on our understanding on the original algorithm appears to little statistical gain compared with ERM. However, after fixing the weights of last two layers of FTNA which gives extra advantages, we achieved similar performances as claimed in the FTNA paper.}.
\end{enumerate}

\begin{figure*}[!ht]
\begin{center}
\setlength\tabcolsep{0pt}
\begin{tabular}{ccccc}
\includegraphics[width=0.4\columnwidth]{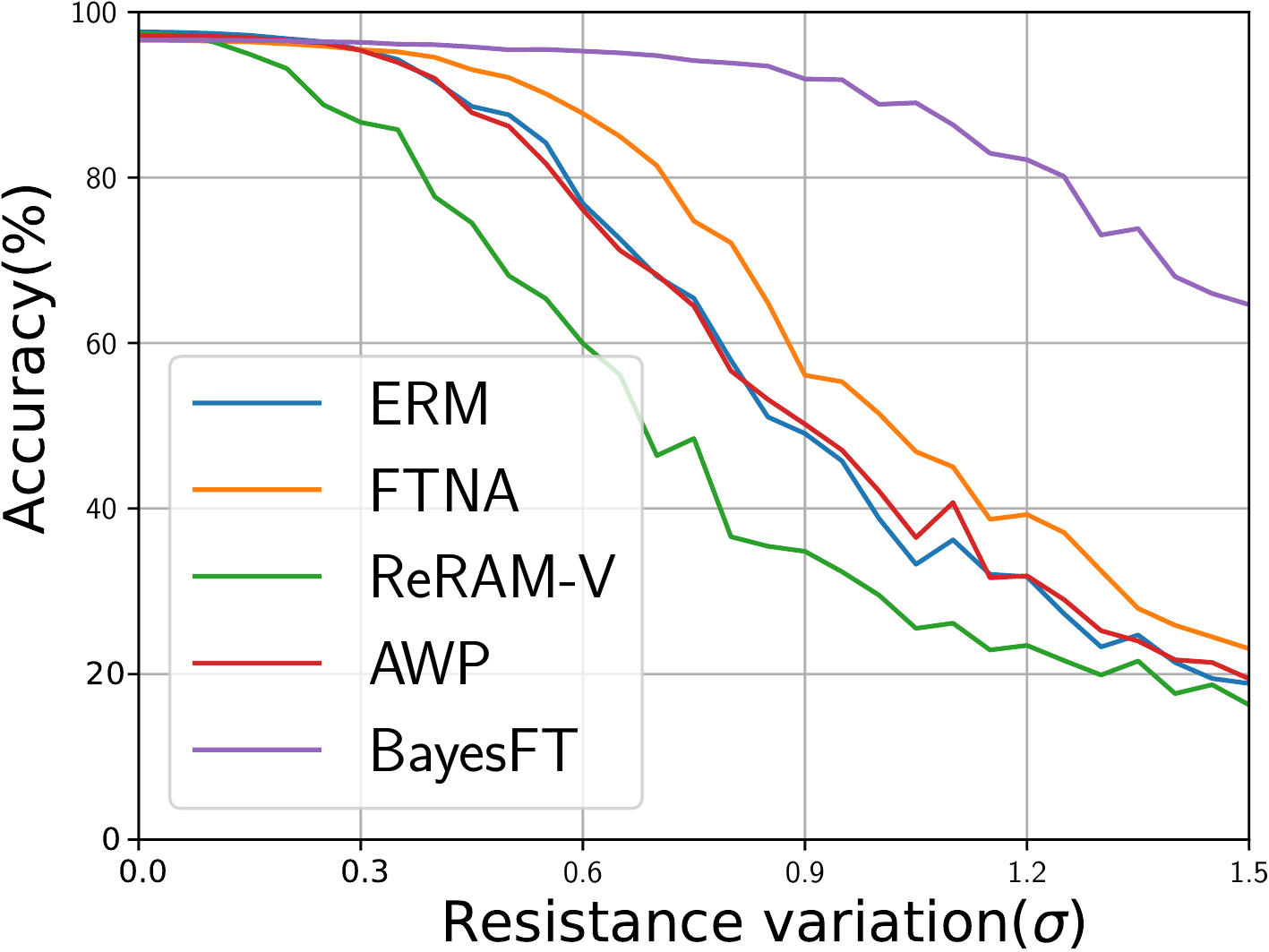}&
\includegraphics[width=0.4\columnwidth]{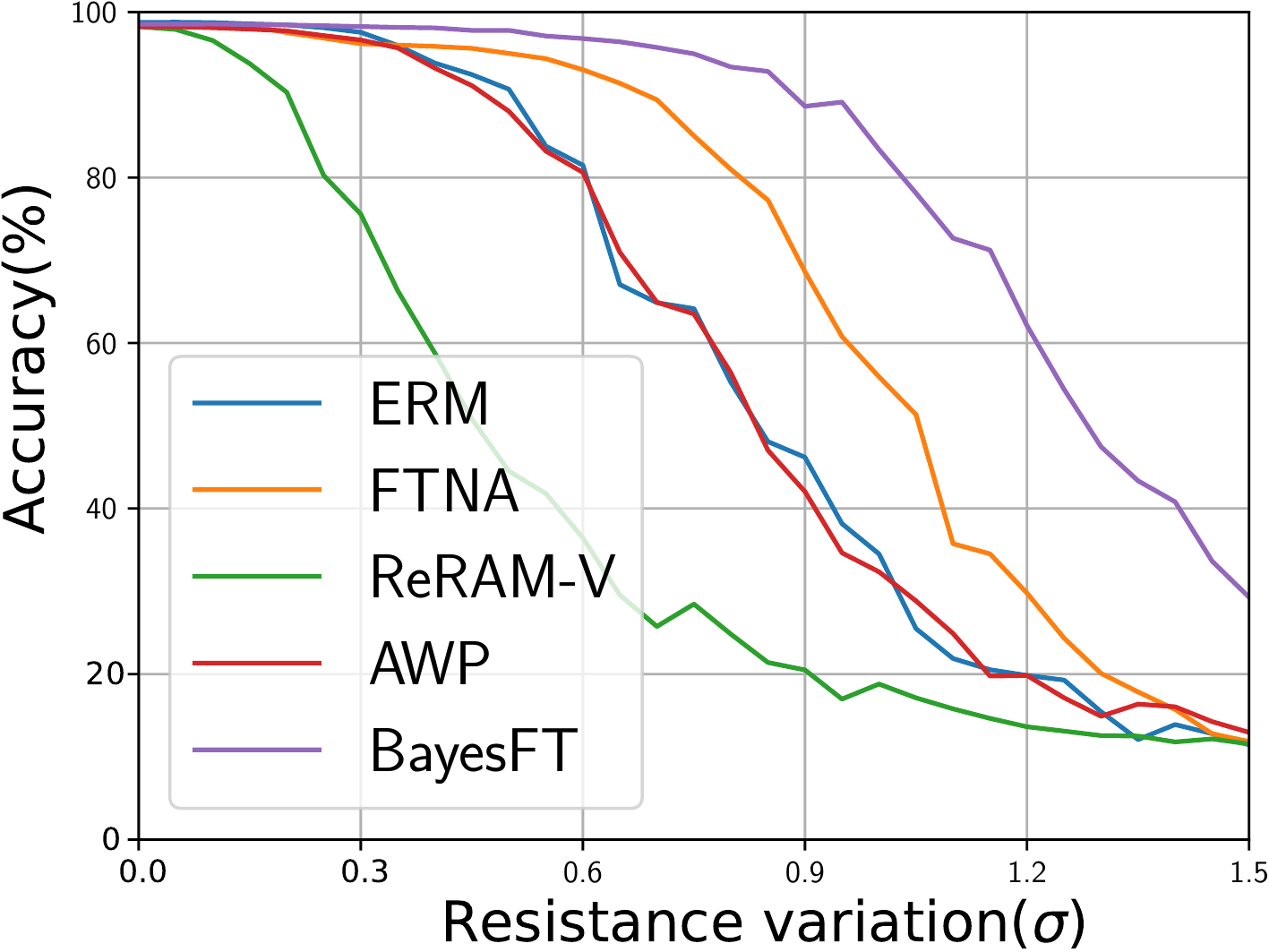}&
\includegraphics[width=0.4\columnwidth]{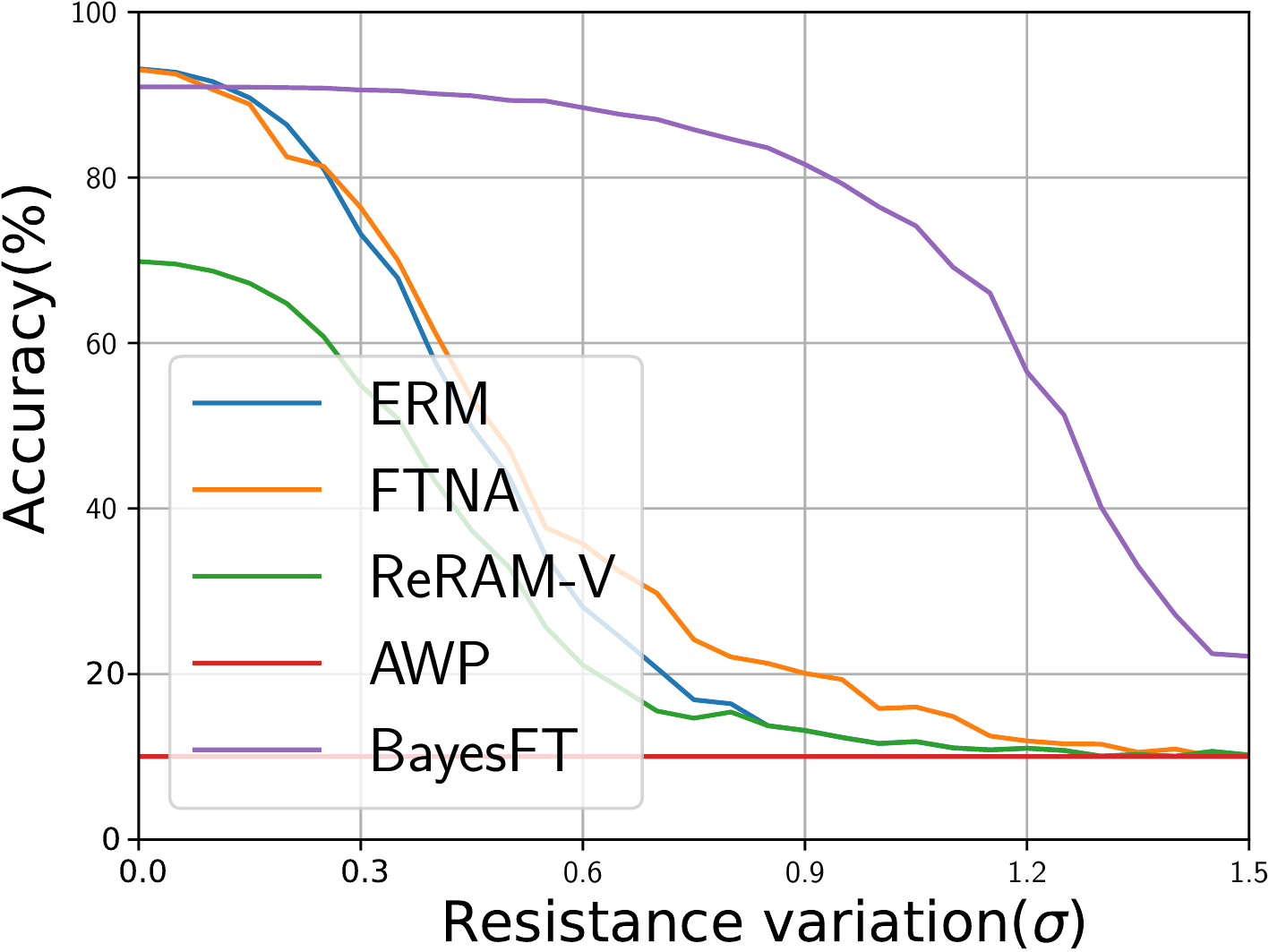}&
\includegraphics[width=0.4\columnwidth]{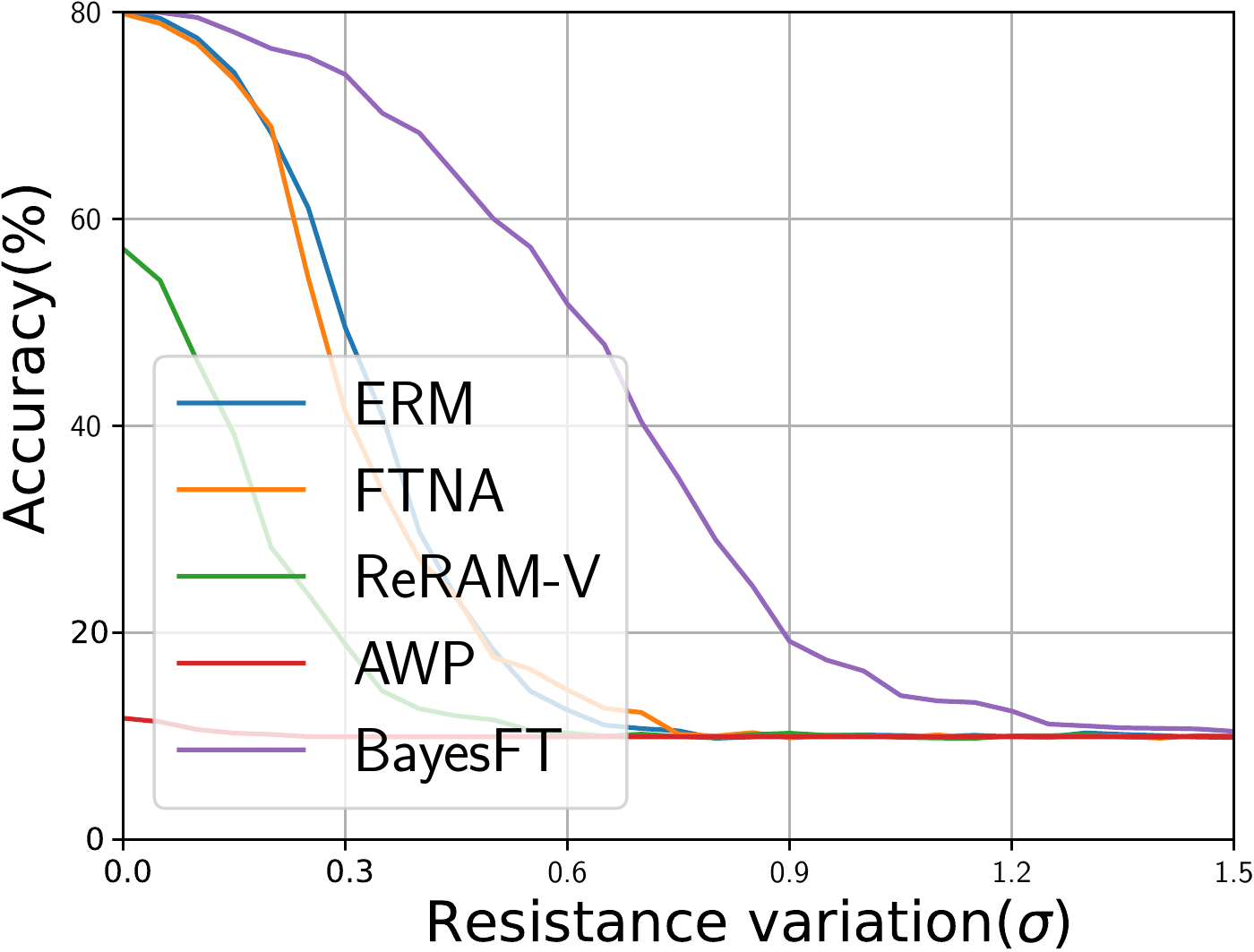}&
\includegraphics[width=0.4\columnwidth]{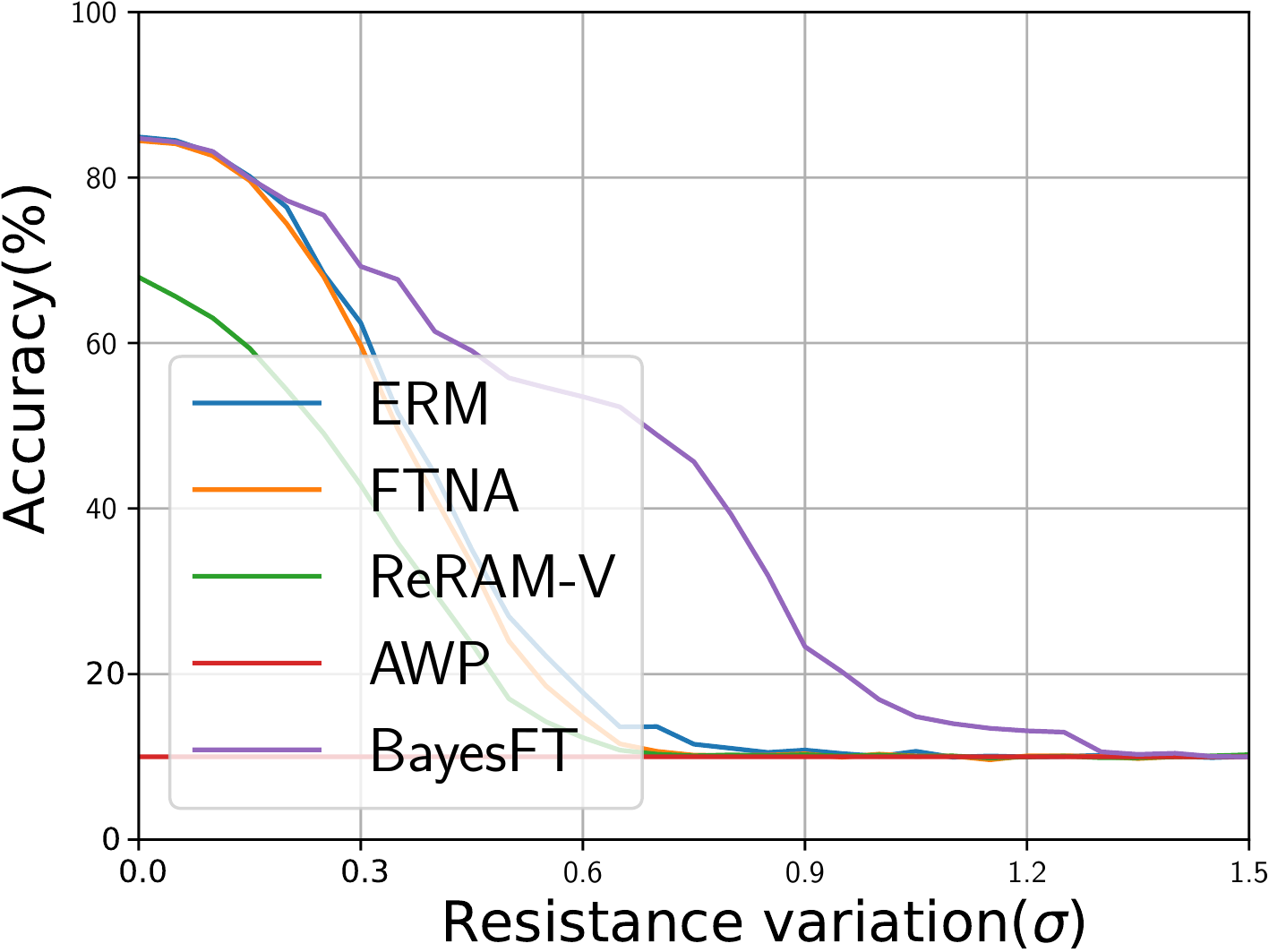}\\
(a) MLP, Mnist &(b) LeNet, Mnist &(c) AlexNet, Cifar10  &(d) ResNet-18, Cifar10 &(e) VGG11, Cifar10 \\

\includegraphics[width=0.4\columnwidth]{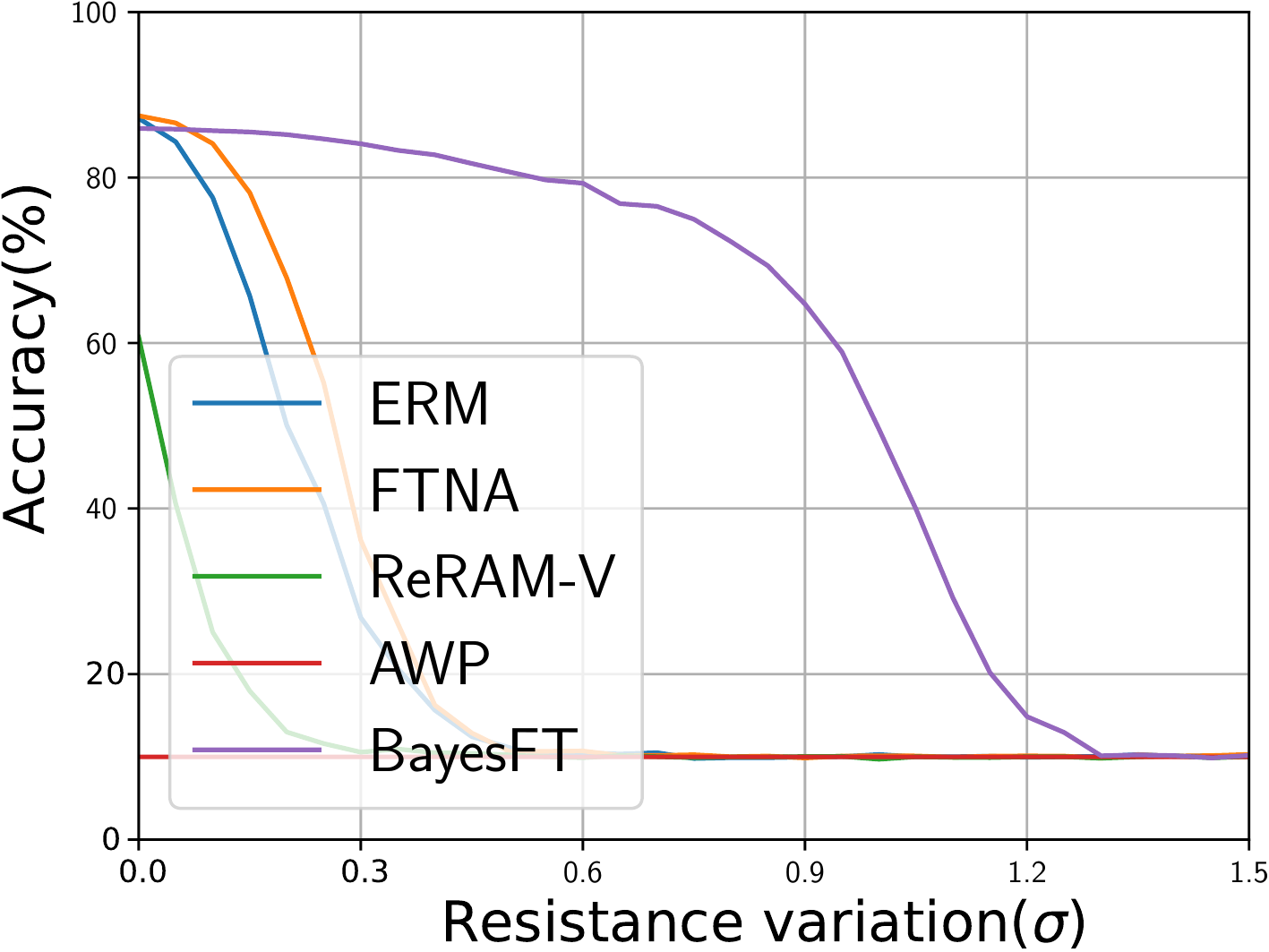}&
\includegraphics[width=0.4\columnwidth]{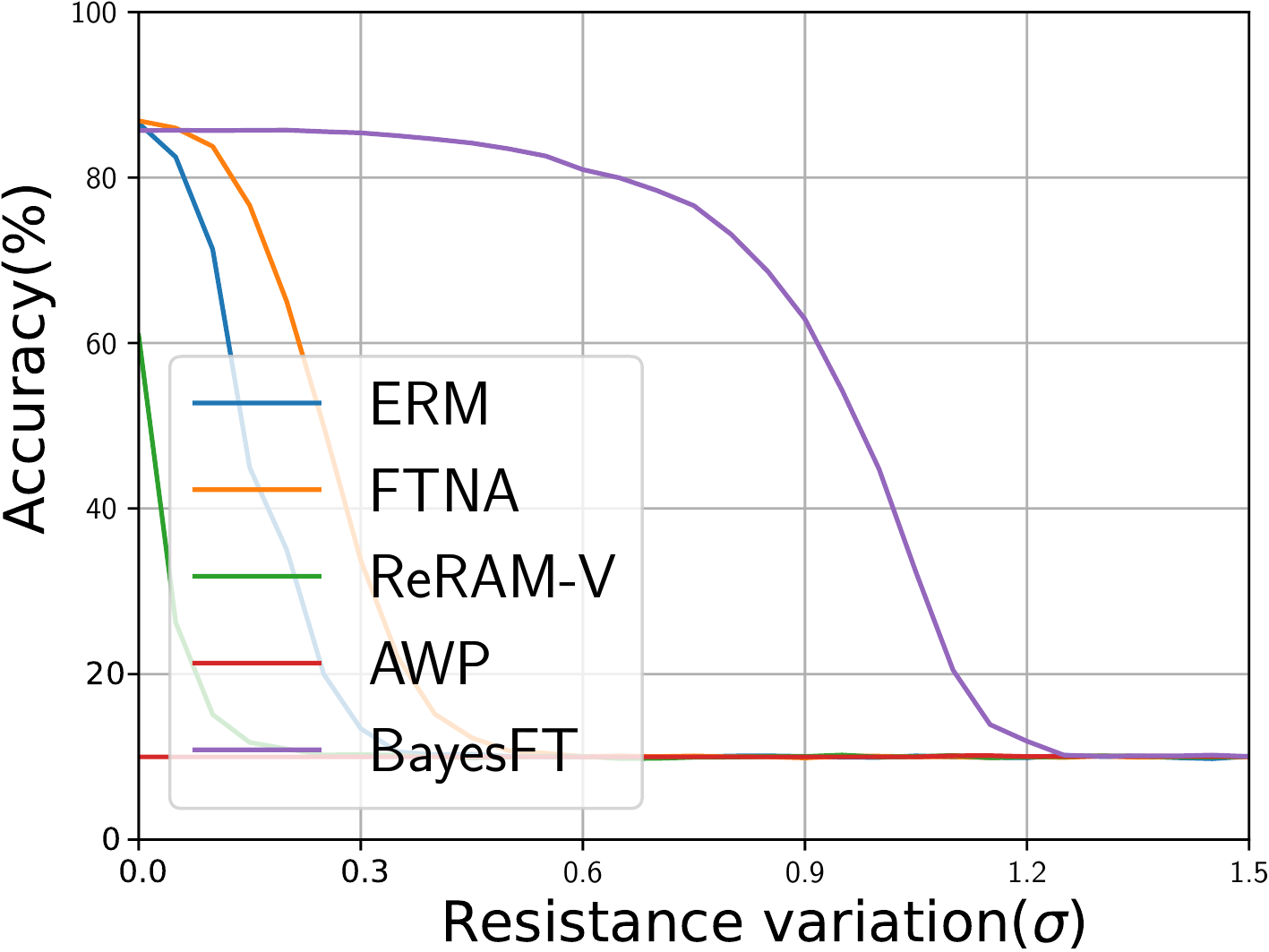}&
\includegraphics[width=0.4\columnwidth]{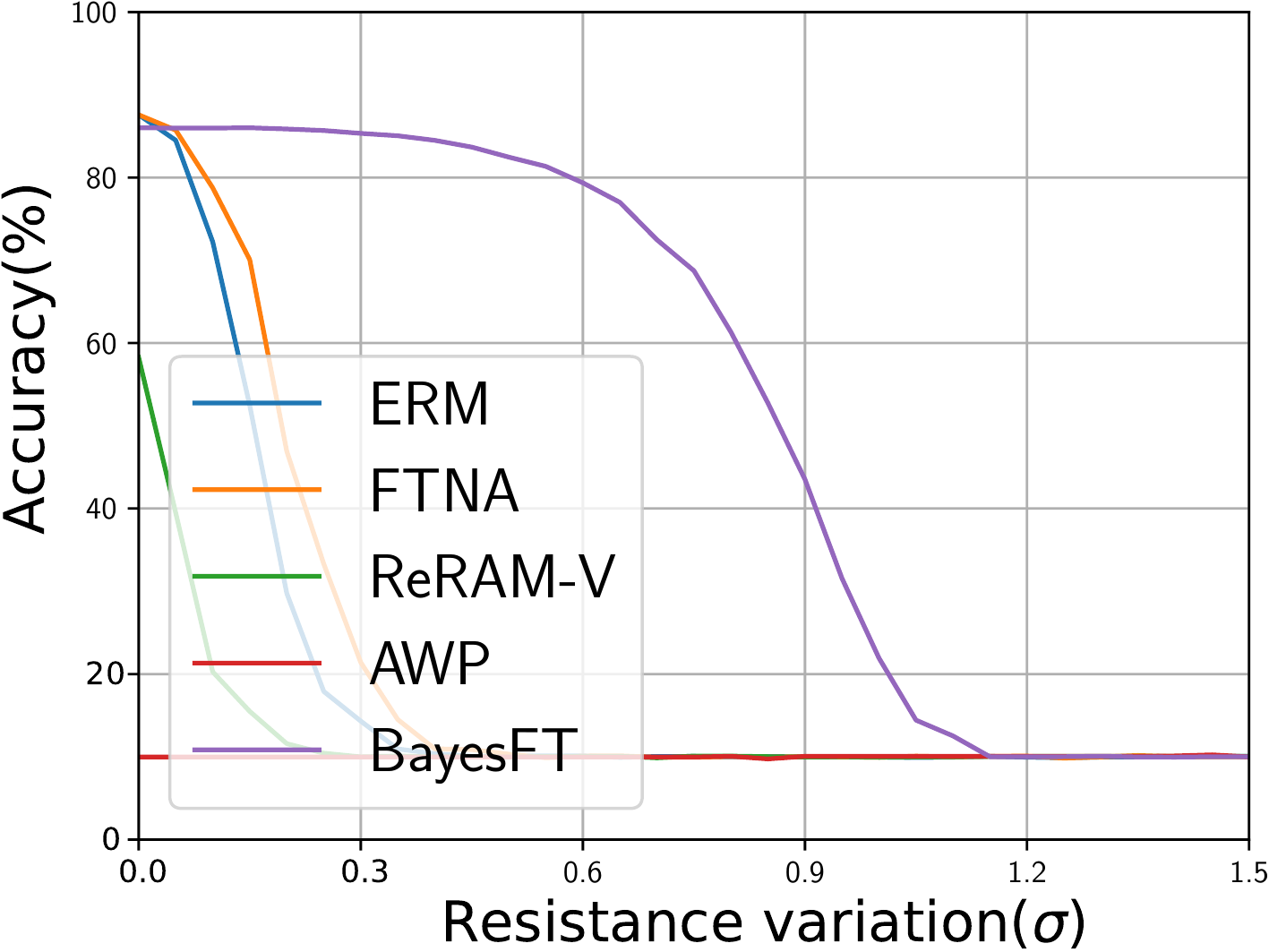}&
\includegraphics[width=0.4\columnwidth]{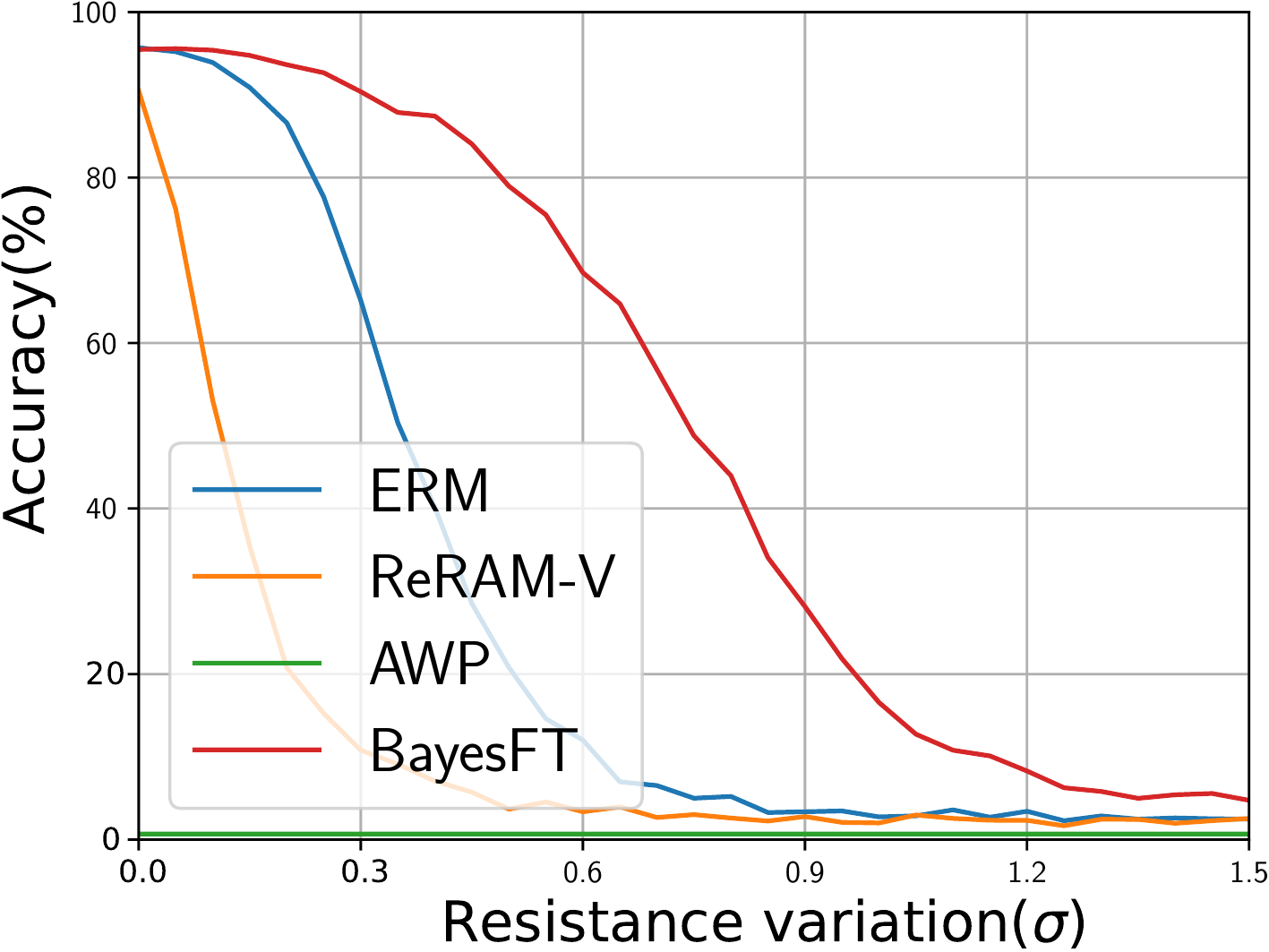}&
\includegraphics[width=0.4\columnwidth]{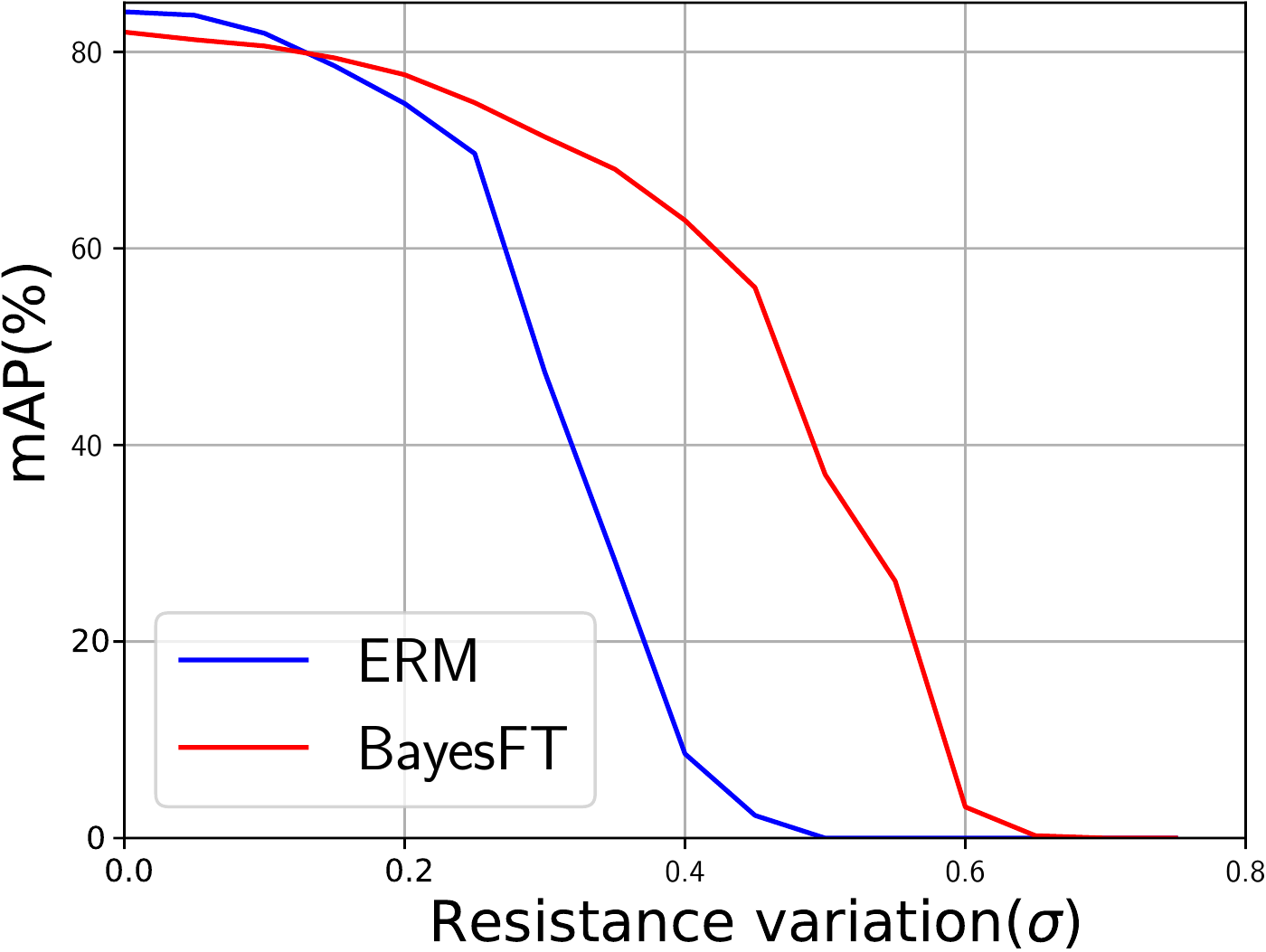}\\
(f) PreAct-18, Cifar10 &(g) PreAct-50, Cifar10 &(h) PreAct-152, Cifar10 &(i) MLP, GTSRB &(j) Object Detection,\\ & & & &PennFudan \\

\end{tabular}
\end{center}
\caption{Numerical experiment results. Better viewed in the zoom-in mode.}
\label{fig:experimentresults}
\end{figure*}

\subsection{Evaluation tasks}
\textbf{Image classification}
We consider MNIST~\cite{lecun-mnisthandwrittendigit-2010}, CIFAR-10~\cite{Krizhevsky09learningmultiple} , and  German Traffic Sign Benchmarks (GTSRB,\cite{Stallkamp-IJCNN-2011}) dataset for image classification tasks. Both MNIST and CIFAR-10 have 10 classes and GTSRB has 43 classes. For MNIST dataset, experiments were carried out on 3-layer MLP and LeNet5\cite{simard2003best}. For CIFAR-10, we provide experimental results on commonly-used larger networks, such as ResNet\cite{he2016deep}, VGG\cite{simonyan2014deep}, AlexNet\cite{Krizhevsky2012ImageNet}, Pre-Activation-ResNet\cite{he2016identity}. For traffic sign recognition, we used the  spatial transformer network that could learn a spatial transformation to automatically transform input images for classification \cite{arcos2018deep}. 

\textbf{Object detection}
PennFudanPed dataset~\cite{Wang07objectdetection} was considered for object detection. The task is to detect pedestrians from input images. For this task, we used the Mask-RCNN network~\cite{he2018mask}. In this task, we found that there is no direct way to implement ReRAM-V, AWP and FTNA methods. We thus compared ERM and BayesFT.

\subsection{Evaluation results}
Figure~\ref{fig:experimentresults} compares the accuracy with baseline algorithms and BayesFT across different levels of resistance variance ($0 \le \sigma \le 1.5$) on 10 different tasks and models. 

\begin{figure*}[!ht]
\begin{center}
\setlength\tabcolsep{0pt}
\begin{tabular}{ccc}
\includegraphics[width=0.64\columnwidth]{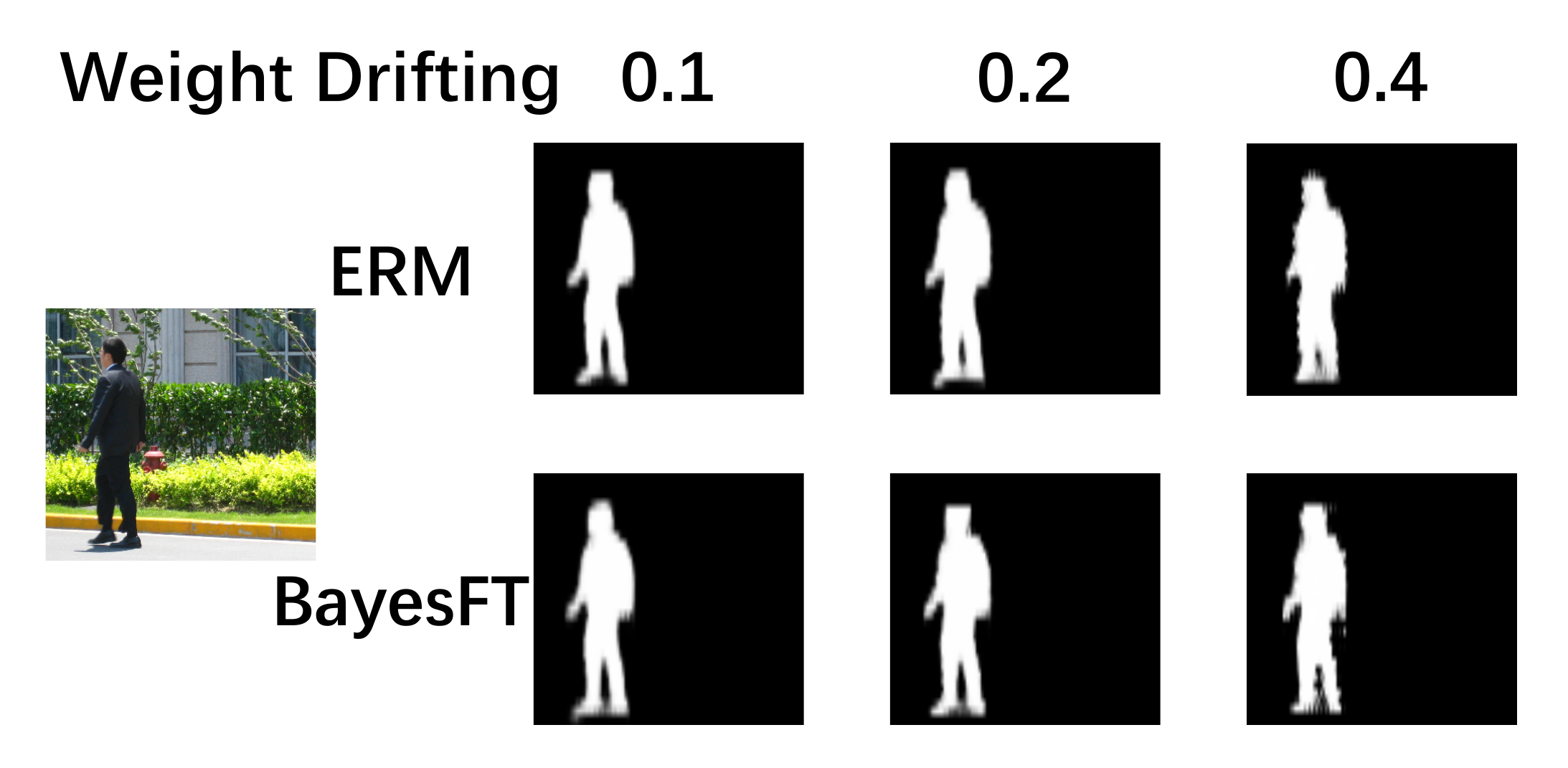} &
\includegraphics[width=0.64\columnwidth]{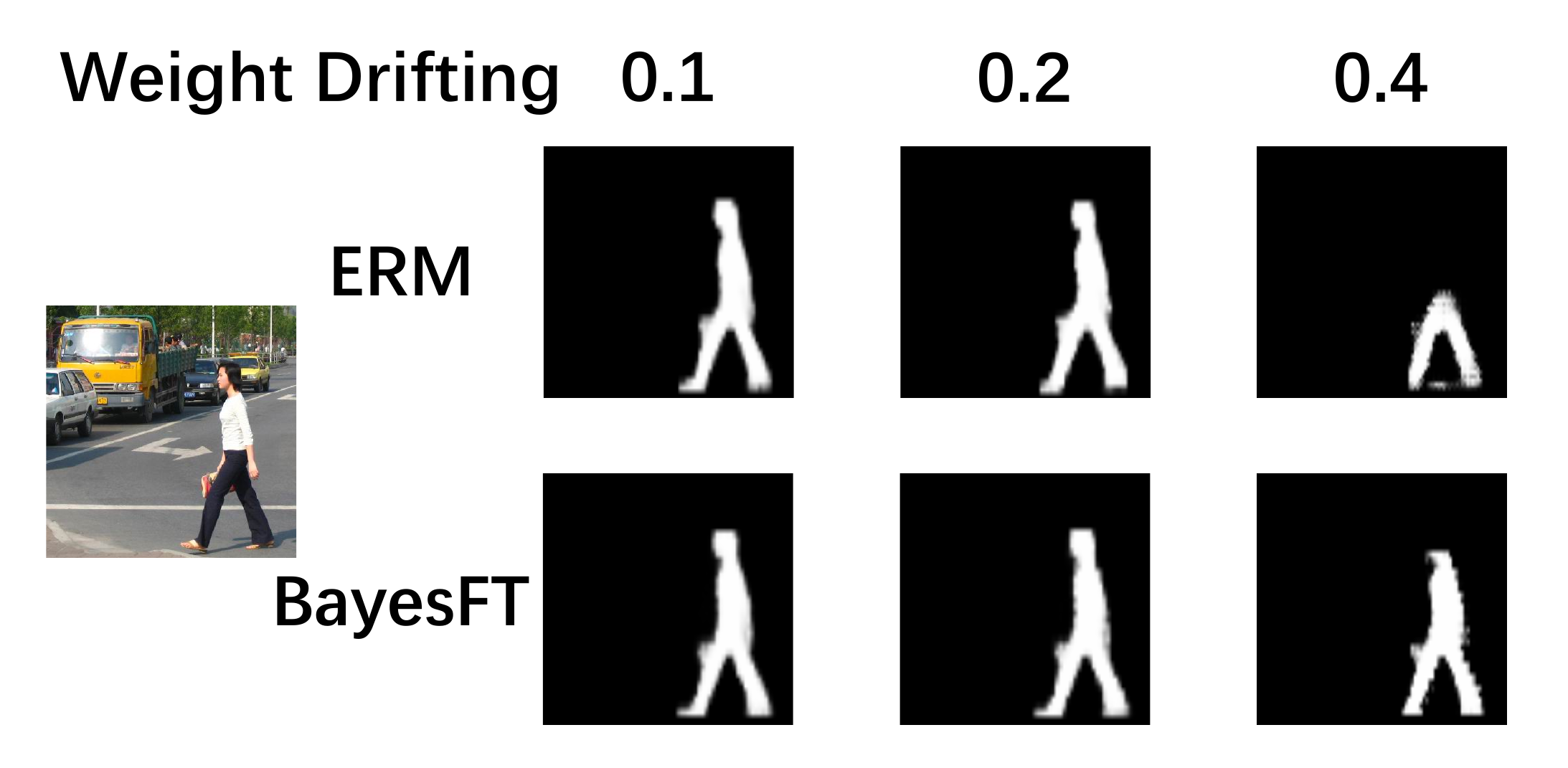} &
\includegraphics[width=0.64\columnwidth]{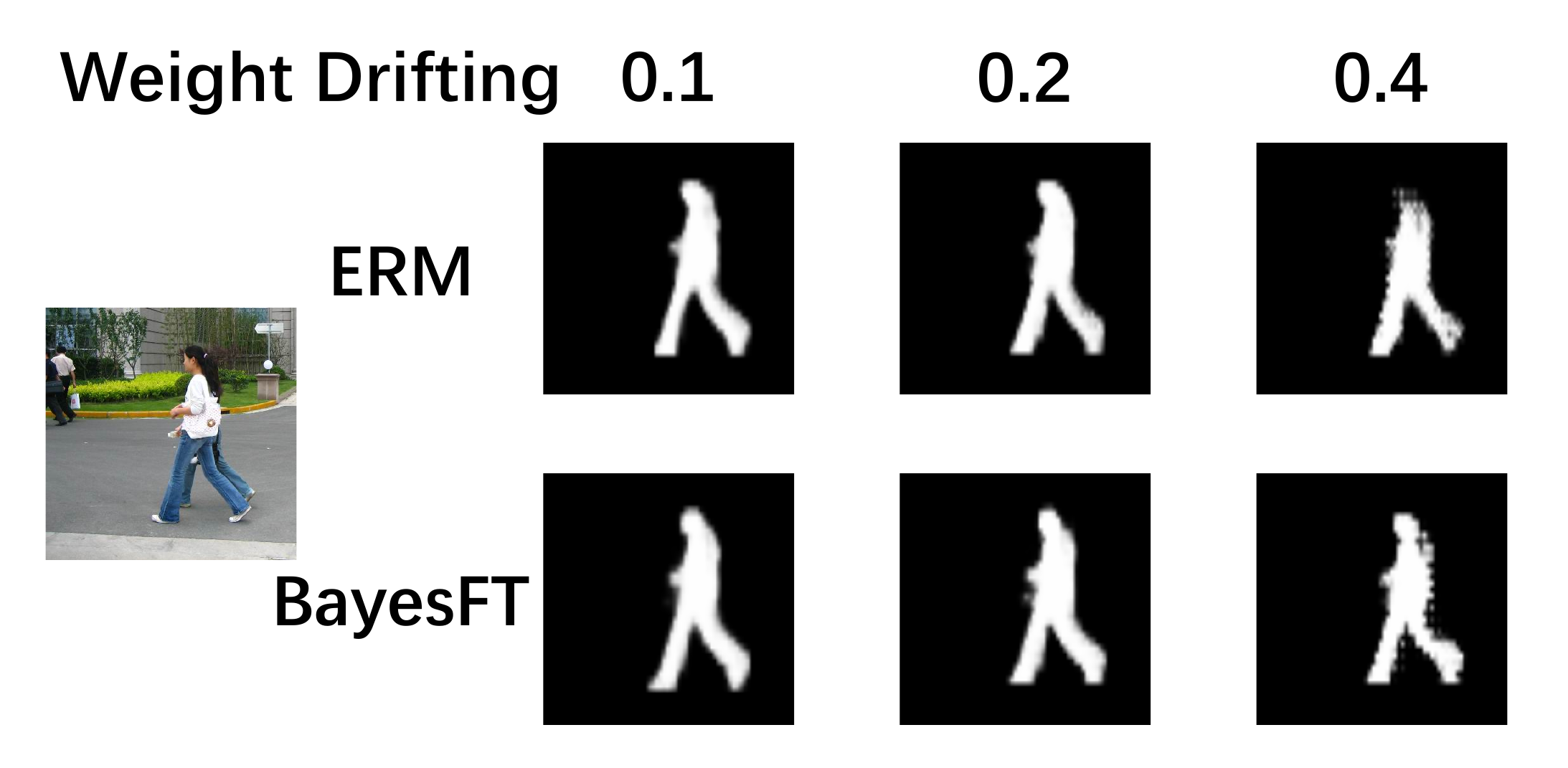} \\
\end{tabular}
\end{center}
\caption{Visualization of object detection.}
\label{fig:visualization_objectdetection}
\end{figure*}
\textbf{MNIST classification}
 On MNIST dataset, all the algorithms except BayesFT experienced severe accuracy degradation as resistance variation increases. As shown in Figure~\ref{fig:experimentresults}(a), BayesFT achieved the best performance among all methods. In the small variance region ($\sigma \le 0.6$), the MLP-MNIST remained perfectly robust with no reduction in classification accuracy. Only when $\sigma$ exceeded 0.9, the accuracy drop slightly. In contrast, the accuracy of other algorithms started to drop significantly when $\sigma$ reaches around 0.2. For larger variance ($1.2 < \sigma < 1.5$), BayesFT's accuracy remained around $70\%$, making it $50\%$ more accurate than other methods. For LeNet\cite{simard2003best} convolutional nerual network result shown in Figure~\ref{fig:experimentresults}(b), BayesFT still outperformed all the other baseline methods. For MNIST dataset, AWP had a similar performance compared to ERM, while ReRAM-V showed an unsatisfactory performance under our experimental setting. Additionally, FTNA boosted the accuracy slightly.

\textbf{CIFAR-10 classification}
The results for CIFAR-10 classification are shown in Figure~\ref{fig:experimentresults}(c)-(g). The baseline methods' accuracy dropped rapidly as the resistance variance increased while BayesFT could still performed stably. For example, in Figure~\ref{fig:experimentresults}(c), the accuracy of BayesFT remained almost the same for $\sigma < 0.6$ and remained greater than $80\%$ for $\sigma < 0.9$.
It is also worth noting that in BayesFT, the accuracy drop was within $0.3\%$ when resistance variance was smaller than 0.3. Similarly, as $\sigma$ increased to 0.6 the accuracy drop was less than $2.5\%$. The improvement of accuracy gained by BayeFT is from $17\%$ to $68\%$ compared to ERM when $\sigma$ varies from $0.3$ to $0.9$. Moreover, BayesFT also demonstrated competitive results on ResNet. Figure~\ref{fig:experimentresults}(f) shows the experiment results for PreAct-18. The ERM robustness is comparably poor to AlexNet and VGG. However, applying BayesFT boosted the robustness significantly with around $50\%$ more accurate classification at variation $0.3 < \sigma < 1$. In addition, by comparing Figure~\ref{fig:experimentresults}(f),(g),(h), an increasingly steeper fall could be observed. This verified the conclusion we drew that the increase in depth of the neural network hurt the robustness to weight drifting. For other algorithms, the performance on CIFAR-10 dataset was not comparable to BayesFT. AWP performed the worst among all algorithms since the strong adversarial attack on the neural network parameters caused training failures.

\textbf{Trafficsign classification}
The BayesFT method boosts performance on this 43-class and randomized input shape classification task with transformer network. ERM performs poorly on this dataset. The accuracy falls quickly and drops to around $10\%$ when $\sigma = 0.6$. In contrast, with the help of BayesFT, the accuracy remains around a tolerable $70\%$. Additionally, for lower sigma variation, say $0< \sigma < 0.4$, the accuracy surpasses $80\%$. 
%FTNA in the original paper only provides the setting for \YE{uniformly}uniform distributed and uniform number of classes dataset so that FTNA \YE{method}methods was not applied here. \YE{Why 'therefore'? I cannot see any causality here}Therefore, BayesFT could be widely applied to different tasks with different type of neural network. 

\textbf{Object detection}
The results for object detection are shown in Figure~\ref{fig:experimentresults}(j). Compared with ERM, BayesFT largely improved the object detection accuracy---mean average precision (mAP). There is also a visualization of object detection results in Figure~\ref{fig:visualization_objectdetection}, from which we can see that ERM lost more details when weight drifting is larger. This example further shows that BayesFT can be seaminglessly applied to other tasks in addition to image classification.

%\begin{figure}[htbp]
  %  \centering
 %   \includegraphics[width=\linewidth]{ObjectDetecti%on-Results/ObjectDetection_Baseline.png}
%    \caption{Object Detection result}
%    \label{fig:Object Detection Results}
%\end{figure}

%\subsubsection{Natural language processing: language modelling}

%\subsubsection{Deep reinforcement learning}

\section{Conclusion and future work}
In this paper, we proposed a novel Bayesian optimization algorithmic framework to search for fault tolerant neural architectures that are robust to memristance drifting in ReRAM DNNs. Extensive numerical experiment results demonstrate the superiority of our algorithm over other baseline methods by a large margin. 

\section*{Acknowledgements}
Nanyang Ye was supported in part by National Key R\&D Program of China  2017YFB1003000, in part by National Natural Science Foundation of China under Grant (No. 61672342, 61671478, 61532012, 61822206, 61832013,  61960206002, 62041205). Ye and Mei are joint corresponding authors.

{\small
\bibliographystyle{IEEEtran}
\bibliography{bayesft.bib}
}

\end{document}